\documentclass{article}

\PassOptionsToPackage{numbers, compress}{natbib}
\usepackage[preprint]{neurips_2026}

\usepackage[utf8]{inputenc}
\usepackage[T1]{fontenc}
\usepackage{hyperref}
\usepackage{url}
\usepackage{booktabs}
\usepackage{amsfonts,amsmath,amssymb}
\usepackage{nicefrac}
\usepackage{microtype}
\usepackage{graphicx}
\usepackage{xcolor}
\usepackage{algorithm}
\usepackage{algorithmic}
\usepackage{pgf}
\usepackage{import}

\definecolor{colorI}{HTML}{1F77B4}   
\definecolor{colorP}{HTML}{D62728}   
\definecolor{colorHt}{HTML}{2CA02C}  
\definecolor{colorElef}{HTML}{7F7F7F}
\newcommand{\colI}[1]{\textcolor{colorI}{#1}}
\newcommand{\colP}[1]{\textcolor{colorP}{#1}}
\newcommand{\colH}[1]{\textcolor{colorHt}{#1}}

\title{On Basis Function Selection for Sparse Gaussian Process Regression}

\author{%
  Marnix Van Soom \\
  AI Lab \\
  Vrije Universiteit Brussel \\
  \texttt{marnix@ai.vub.ac.be} \\
  \And
  Ivan De Boi \\
  InViLab \\
  University of Antwerp \\
  \texttt{ivan.deboi@uantwerpen.be} \\
}

\begin{document}

\maketitle

\begin{abstract}
Sparse Gaussian processes achieve $\mathcal{O}(N)$ inference by replacing the kernel with an appropriate expansion in a fixed basis $\{\phi_j\}$ on the input space.
Given a compute budget $M \ll N$, practitioners conventionally \emph{truncate} the basis to its first $M$ entries.
Nothing in the formalism, however, prevents one from \emph{selecting} only those $M$ basis functions that matter for the data at hand.
This would avoid spending budget on basis functions where there is no signal, but it requires a criterion for ranking the candidates.
We propose three such criteria derived from an information-theoretic view of the basis-function selection problem.
Each criterion matches a different state of knowledge at selection time: (no data), (no prior), and an (in-between) state.
We then study the performance of truncation vs. selection strategies on six UCI regression benchmarks across three basis families: Hilbert-space Gaussian processes (HSGP), variational Fourier features (VFF), and variational inducing spherical harmonics (VISH).
We observe that the (no data) criterion is a safe default, matching or improving on truncation for HSGP, VFF, and VISH, with substantial gains for VISH and improvements over a recently developed selection heuristic for that basis family.
The data-aware (no-prior) and (in-between) criteria provide substantial gains over truncation specifically for HSGP, which is the most broadly used of the three families in practice.
\end{abstract}

\section{Introduction}
\label{sec:intro}

Gaussian processes \citep{Rasmussen2006} are a powerful class of models for regression.
Given $N$ noisy observations $\mathbf{y} = (y_1, \ldots, y_N)$ at inputs $X = (x_1, \ldots, x_N) \in \mathbb{R}^{N \times D}$, one models the response through a latent function $f$ with the prior $f \sim \mathcal{GP}(0, k_\theta)$, and learns the kernel hyperparameters $\theta$ from the data.
Vanilla inference is, however, $\mathcal{O}(N^3)$ in time because of the kernel-matrix solve, which is prohibitive for the moderately large $N$ encountered in practice.

Sparse GPs \citep{Leibfried2022} avoid this cost by replacing the kernel with an expansion in a basis $\{\phi_j\}$ on the input space, informally written as
\begin{equation}
  f(x) \;=\; \sum_{j=1}^{\infty} w_j \,\phi_j(x), \qquad w \sim \mathcal{N}\!\big(0, \Sigma(\theta)\big).
  \label{eq:expansion}
\end{equation}
The sparse-GP method fixes both the basis $\{\phi_j\}$ and the covariance $\Sigma(\theta)$ on the coefficients, and typically also requires additional user choices, such as the input domain on which the GP is approximated.
One such choice stands out: the number $M \ll N$ of basis functions actually used for inference.
Computing the marginal likelihood and the predictive mean scales as $\mathcal{O}(NM^2 + M^p)$, with $p$ depending on the method, so $M$ is essentially the user's compute budget.
The conventional way to spend that budget is to \emph{truncate} the expansion in Equation~\eqref{eq:expansion} to its first $M$ entries, in some ordering of the index $j$.

For example, take a regression problem in $D = 2$ input dimensions and apply the Hilbert-space sparse-GP method, which is the most common of the three methods considered here (see Section~\ref{sec:bg} for the construction).
The expansion in Equation~\eqref{eq:expansion} then takes the form
\begin{equation}
  f(x_1, x_2) \;=\; \sum_{j_1 = 1}^{\infty} \sum_{j_2 = 1}^{\infty} w_{j_1, j_2}\, \phi_{j_1, j_2}(x_1, x_2),
  \qquad
  w_{j_1, j_2} \sim \mathcal{N}\!\big(0, \lambda_{j_1, j_2}(\theta)\big),
  \label{eq:hsgp-d2}
\end{equation}
where the basis function $\phi_{j_1, j_2}$ is a product of two one-dimensional sinusoids, indexed by per-axis frequencies $(j_1, j_2)$, and the prior variance $\lambda_{j_1, j_2}(\theta)$ is the kernel's spectral density evaluated at the frequency corresponding to $(j_1, j_2)$.
Suppose now that our compute budget allows only $M = 10$ basis functions out of the infinite expansion: which ten pairs $(j_1, j_2)$ should we keep, that is, which frequencies should we model?

\begin{figure}[tbp]
  \centering
  \def\panelw{0.20\linewidth}%
  \def\opw{0.062\linewidth}%
  \newcommand{\panelone}{%
    \begin{minipage}[c]{\panelw}\centering
      \footnotesize $M_d = (5, 2)$\\[1pt]
      \includegraphics[width=\linewidth]{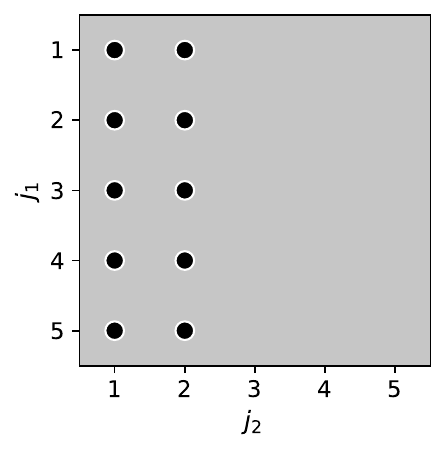}\\
      \footnotesize $\text{NLL} = 0.135$\\[3pt]
      \footnotesize\textbf{truncation}
    \end{minipage}}%
  \newcommand{\paneltwo}{%
    \begin{minipage}[c]{\panelw}\centering
      \footnotesize $\colI{I_j} = \lambda_j$\\[1pt]
      \includegraphics[width=\linewidth]{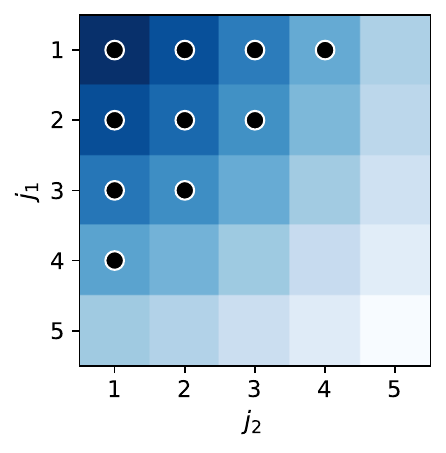}\\
      \footnotesize $\text{NLL} = 0.095$\\[3pt]
      \footnotesize\textbf{(no data)}
    \end{minipage}}%
  \newcommand{\panelthree}{%
    \begin{minipage}[c]{\panelw}\centering
      \footnotesize $\colP{P_j} = |a_j|^2$\\[1pt]
      \includegraphics[width=\linewidth]{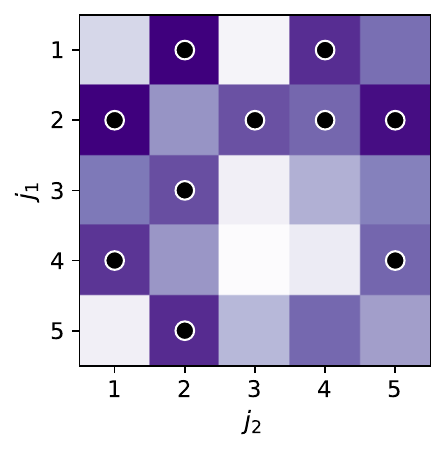}\\
      \footnotesize $\text{NLL} = 0.127$\\[3pt]
      \footnotesize\textbf{(no prior)}
    \end{minipage}}%
  \newcommand{\panelfour}{%
    \begin{minipage}[c]{\panelw}\centering
      \footnotesize $\colH{\tilde H_j} = \lambda_j |a_j|^2$\\[1pt]
      \includegraphics[width=\linewidth]{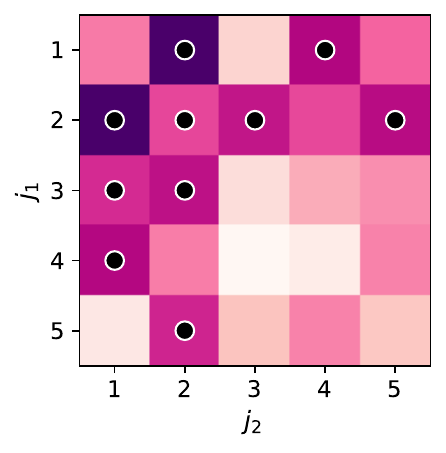}\\
      \footnotesize $\boldsymbol{\text{NLL} = 0.085}$\\[3pt]
      \footnotesize\textbf{(in-between)}
    \end{minipage}}%
  \newcommand{\opbox}[1]{%
    \begin{minipage}[c]{\opw}\centering{\Large$#1$}\end{minipage}}%
  \panelone\hspace{0.018\linewidth}%
  \colorbox{black!7}{%
    \paneltwo\opbox{\times}\panelthree\opbox{=}\panelfour
  }
  \caption{The four basis-function selection strategies we consider in the paper, illustrated on a toy dataset.
  From left to right: truncation $M_d = (5, 2)$, the (no data) criterion $\colI{I_j} = \lambda_j$, the (no prior) criterion $\colP{P_j} = |a_j|^2$, and the (in-between) criterion $\colH{\tilde H_j} = \lambda_j |a_j|^2$, computed as the pointwise product of the previous two.
  Selected basis functions are marked by black dots, and the test NLL is reported underneath each panel.
  Heatmap colours indicate the strength of the criterion (e.g.\ $\colI{I_{j_1, j_2}}$ in panel 2) at each grid point.
  On this example $\colH{\tilde H_j}$ performs best by combining the low-frequency basis functions favoured by $\colI{I_j}$ with the higher-energy ones that $\colP{P_j}$ alone prefers.}
  \label{fig:motivational}
\end{figure}

Figure~\ref{fig:motivational} (left panel) shows one such allocation on a toy dataset: the choice $M_d = (5, 2)$ arranges its basis functions as a rectangle in the $(j_1, j_2)$ grid, with the test negative log-likelihood (NLL) reported underneath the panel.
Such a rectangular allocation is the convention in available HSGP implementations: the user picks a count $M_d$ per axis,\footnote{For example, PyMC's \texttt{pymc.gp.HSGP(m=[$M_1, \ldots, M_D$])} and NumPyro's \texttt{hsgp\_squared\_exponential(m=[\ldots])} both take a per-dimension list \citep{Solin2020, Riutort-Mayol2020}, while the brms R package exposes a single $k$ across all axes via \texttt{gp(\ldots, k=\ldots)}.} and the method then keeps every basis function with $j_d \le M_d$, giving $M = \prod_{d=1}^{D} M_d$ basis functions in total.
This rule, however, is not unique even at fixed $M$: a budget of $M = 10$ admits the allocations $(5, 2)$, $(2, 5)$, $(10, 1)$, and any other partition with the same product, and the user is left to pick by heuristics and trial and error.

This non-uniqueness in truncation points to a deeper fact: nothing in the sparse-GP formalism actually requires us to truncate sequentially at all.
We can instead rank the candidates by any other criterion and keep the top $M$ by that criterion.
We call this \emph{selection}, in contrast with the \emph{truncation} baseline.
The question is then which criterion to choose, and whether a cheap one can offer gains over truncation.

This paper proposes three such criteria, derived from an information-theoretic view of the problem.
They are shown in the grey panels of Figure~\ref{fig:motivational} alongside the truncation baseline.
Each criterion produces a scalar value (the score) at every candidate index $j = (j_1, j_2)$, and selection keeps the top $M$ candidates by that value.

We show in Section~\ref{sec:score} that the criteria correspond to different states of knowledge at selection time.
The (no data) criterion $\colI{I_j} = \lambda_j(\theta)$ ranks candidates by the kernel-prior weight alone (panel 2).
The (no prior) criterion $\colP{P_j} = |a_j|^2$ ranks by the squared data projection $a_j := \phi_j(X)^\top y$ alone (panel 3).
The (in-between) criterion $\colH{\tilde H_j} = \lambda_j |a_j|^2$ uses both factors (panel 4).
Black dots mark the selected top-$M = 10$ basis functions, and each criterion picks a different ten.
The resulting fit, and therefore the test NLL reported under each panel, varies with the choice of criterion.

HSGP truncation can only carve rectangular subsets out of the candidate grid.
The three selection criteria are not bound by that shape.
Even without looking at the data, the (no data) criterion $\colI{I_j}$ already departs from the rectangle and improves substantially on truncation.
On this toy example, the (in-between) criterion $\colH{\tilde H_j}$ does best, notably better than the rectangular truncation.

We shall see that this pattern does not hold in general.
We test whether selection improves over truncation on six UCI regression benchmarks across three sparse-GP methods: HSGP, variational Fourier features (VFF), and variational inducing spherical harmonics (VISH).
The three methods are different but related, each with its own basis $\phi_j$ and prior variance $\lambda_j(\theta)$.
Ranking by the (no data) criterion $\colI{I_j}$ is a safe default that matches or improves on truncation for all three methods, with the largest gains on VISH where it also outperforms the recently proposed phase-truncation heuristic of \citet{Eleftheriadis2023}.
The data-aware (no prior) criterion $\colP{P_j}$ and (in-between) criterion $\colH{\tilde H_j}$ offer substantial gains over truncation specifically for HSGP, the most broadly used of the three methods in practice.

The rest of the paper develops these ideas.
Sections~\ref{sec:bg} and~\ref{sec:score} fix notation and derive the three criteria.
Sections~\ref{sec:hsgp}--\ref{sec:vish} report the per-method experiments.
Section~\ref{sec:related} places the criteria in context, and Section~\ref{sec:disc} discusses what we learned.

\section{Background}
\label{sec:bg}

We consider the standard Gaussian process regression setting: a Gaussian process $f \sim \mathcal{GP}(0, k_\theta)$ on an input space $\mathcal{X} \ni x = (x_1, \ldots, x_D)$ observed at $N$ inputs $X = (x_1, \ldots, x_N)^\top$ with additive Gaussian noise $y_i = f(x_i) + \varepsilon_i$, $\varepsilon_i \sim \mathcal{N}(0, \sigma^2)$.
The prior mean is zero throughout this paper, with any mean shift absorbed by standardising the response in pre-processing.
Exact GP inference scales as $\mathcal{O}(N^3)$ in compute and is infeasible at the dataset sizes we typically care about.
We therefore work with sparse GPs throughout.
The three methods we consider (HSGP, VFF, and VISH) each specify the sparse GP through a basis-function expansion of $f$.

Each method supplies an infinite basis-function expansion of the form Equation~\eqref{eq:expansion}.\footnote{The basis-function expansion is one of two equivalent ways to specify a sparse GP.
The other goes through inducing variables and a variational distribution \citep{Leibfried2022}.
The two views agree at the level of the per-basis-function prior weight $\lambda_j(\theta)$ and the data projection $a_j$ that our criteria depend on. Appendix~\ref{app:setup} discusses the connection.}
The index $j$ is in general a multi-index whose structure is set by the method.
In practice we keep only $M$ of these basis functions, either by truncation (which keeps the first $M$ in the natural ordering of $j$) or by selection (which scores a finite candidate grid $\{\phi_j\}_{j=1}^{J}$ and keeps the top $M$ by score).
The \emph{candidate grid size} $J$ is therefore needed only for selection, and in general $M \ll J \ll N$ when $N$ is large.
The selection problem itself is to choose a subset $\mathcal{M} \subset \{1, \ldots, J\}$ of size $M$, and to fit $\theta$ on that subset.
The three methods specialise the basis and the prior as follows.

\paragraph{Hilbert-space Gaussian processes (HSGP)} \citep{Solin2020, Riutort-Mayol2020} use Dirichlet Laplacian eigenfunctions on a box $[-L, L]^D$, indexed by $j = (j_1, \ldots, j_D)$, with prior variance $\lambda_j(\theta) = S_\theta(\omega_j)$ at the Laplacian frequency vector $\omega_j = (\pi j_1 / 2L, \ldots, \pi j_D / 2L)$, and are the most broadly used of the three methods in practice.
Details in Appendix~\ref{app:hsgp-detail}.

\paragraph{Variational Fourier features (VFF)} \citep{Hensman2017} use an additive model $f(x) = \sum_{d=1}^{D} f_d(x_d)$ where each $f_d$ is expanded in windowed cosine and sine features at per-axis frequencies $j_d$.
Candidates are indexed by $j = (d, j_d)$, with prior variance $\lambda_j(\theta) \approx S_\theta(\omega_{j_d})/T$ the kernel's spectral density at the harmonic frequency on the per-axis box of length $T$.
Details in Appendix~\ref{app:vff-detail}.

\paragraph{Variational inducing spherical harmonics (VISH)} \citep{Dutordoir2020, Eleftheriadis2023} use spherical harmonics on $S^{D}$ for inputs lifted to the unit sphere by $z_n = (x_n, 1)/\|(x_n, 1)\|$, indexed by $j = (\ell, k)$ for degree $\ell$ and orientation $k$, with prior variance $\lambda_j(\theta) = \lambda_\ell(\theta)$ shared across orientations within a degree by Funk--Hecke.
We use the order-$1$ arc-cosine kernel of \citet{Cho2009} throughout.
Details in Appendix~\ref{app:vish-detail}.

For each method we follow the authors' recommended training procedure: the closed-form marginal likelihood for HSGP, and the closed-form collapsed Titsias bound for VFF and VISH.
We reproduce the published truncation baselines as faithfully as possible, with the full experimental protocol in Appendix~\ref{app:expt-uci}.

\section{The selection criteria}
\label{sec:score}

Each candidate index $j$ identifies a basis-function coefficient $u_j$ in the expansion of Equation~\eqref{eq:expansion}, with marginal prior $u_j \sim \mathcal{N}(0, \lambda_j(\theta))$ in all three methods (Appendix~\ref{app:setup}).
We want to know whether $u_j$ is worth keeping in $\mathcal{M}$.
The natural quantity is how much observing the data $y$ changes the belief about $u_j$, measured as the Kullback--Leibler divergence from posterior to prior:
\begin{equation}
  H_j(\theta, y)
  \;:=\;
  D_{\mathrm{KL}}\!\big(p(u_j \mid y)\,\|\,p(u_j)\big).
  \label{eq:KL}
\end{equation}
This is the per-basis-function information gain: the same $D_{\mathrm{KL}}$ objective that \citet{rasmussen2006gaussian} use for placing inducing points and \citet{Seeger2003, krause2008near} use for sensor selection, applied here to the discrete index $j$ that enumerates a fixed candidate basis.
A coefficient $u_j$ with large $H_j$ is one whose belief is shifted strongly by the data, so a large $H_j$ flags a candidate worth keeping.
Since $H_j \ge 0$ we use it directly as a non-negative scoring rule for the basis-function selection problem.

The prior $p(u_j) = \mathcal{N}(0, \lambda_j(\theta))$ is univariate Gaussian by construction, and the marginal posterior $p(u_j \mid y)$ is univariate Gaussian by conjugacy in Bayesian linear regression on the selected basis.
Equation~\eqref{eq:KL} therefore has a closed form.
With the per-basis-function data projection and signal-to-noise ratio
\begin{equation}
  a_j \;:=\; \phi_j(X)^\top y,
  \qquad
  \rho_j \;:=\; \frac{\lambda_j(\theta)\,\|\phi_j(X)\|^2}{\sigma^2},
  \label{eq:aj_rhoj}
\end{equation}
this closed form is
\begin{equation}
  H_j(\theta, y)
  \;=\;
  \tfrac{1}{2}\Big[\log(1 + \rho_j) - \tfrac{\rho_j}{1 + \rho_j}\Big]
  \;+\;
  \frac{\lambda_j(\theta)\,|a_j|^2}{2\,\sigma^4\,(1 + \rho_j)^2},
  \label{eq:H_full}
\end{equation}
under the empirical-decoupling assumption $\Phi^\top \Phi \approx \mathrm{diag}(\|\phi_j(X)\|^2)$, the standard regime in the basis-function case (Appendix~\ref{app:derivation}).
The first bracket is a variance-shrinkage term that is independent of the data.
The second bracket is the data-driven term.
Evaluating Equation~\eqref{eq:H_full} requires both the noise scale $\sigma^2$ and the kernel eigenvalues $\lambda_j(\theta)$, neither of which is fit yet at selection time.

\paragraph{Three limits, three criteria}
Three limits of $H_j$ correspond to three states of knowledge at selection time about how the prior weight and the data interact, and each gives a closed-form ranking criterion on the candidate basis.

\emph{(i) The no-data limit: the eigenvalue criterion $\colI{I_j}$.}
Under the prior, $a_j$ is mean-zero, and a short calculation (Appendix~\ref{app:derivation}) gives the data-averaged information gain
\begin{equation}
  \mathbb{E}_y\!\big[H_j(\theta, y)\big]
  \;=\;
  \tfrac{1}{2}\log(1 + \rho_j),
  \label{eq:I_Ey}
\end{equation}
monotone in $\rho_j$ and therefore in the prior weight $\lambda_j(\theta)$ at fixed $\theta$ and $\sigma^2$.
Ranking by Equation~\eqref{eq:I_Ey} therefore reduces to ranking by
\begin{equation}
  \colI{I_j(\theta)}
  \;:=\;
  \lambda_j(\theta).
  \label{eq:I}
\end{equation}
At fixed $\theta$, $\colI{I_j}$ ranks candidates by prior weight alone.
It is the closest per-basis-function analogue of conventional truncation, since for stationary kernels the spectral density decays radially and $\colI{I_j}$ keeps the lowest-frequency basis functions first.
The two need not agree, however: practitioner-default truncation enforces a structural shape (rectangular cube, harmonic shells, contiguous frequencies), while $\colI{I_j}$ ranks every candidate freely (Figure~\ref{fig:motivational}, panels~1 and~2).

\emph{(ii) The no-prior limit: the data-energy criterion $\colP{P_j}$.}
Dropping the kernel factor in Equation~\eqref{eq:H_full} leaves the data-driven piece, $|\phi_j(X)^\top y|^2 / [2\sigma^4 (1+\rho_j)^2]$.
At fixed $\theta$ and $\sigma^2$ the prefactors are basis-function-independent up to $(1+\rho_j)^2$, which itself is a function of $\lambda_j$.
Ignoring that residual prior dependence gives the data-energy criterion
\begin{equation}
  \colP{P_j(y)}
  \;:=\;
  |\phi_j(X)^\top y|^{2},
  \label{eq:P}
\end{equation}
the squared projection of the data onto $\phi_j$.
A second, more principled, route arrives at the same ranking: if $\sigma^2$ and $\lambda_j$ are jointly estimated from the data and substituted into the full Equation~\eqref{eq:H_full}, the resulting per-basis-function score is rank-equivalent to $\colP{P_j}$ (Appendix~\ref{app:derivation}).
The fully data-driven $H_j$ therefore collapses to the data-only ordering.

\emph{(iii) The weak-observation limit: the criterion $\colH{\tilde H_j}$.}
Selection happens before the kernel hyperparameters are fit, so the practical setting is one of weak observation: the score is evaluated at a starting point $\theta_0$ rather than at a maximum-likelihood estimate.
Taking $\rho_j \to 0$ in Equation~\eqref{eq:H_full},
\begin{equation}
  H_j(\theta, y)
  \;=\;
  \frac{\lambda_j(\theta)\,|a_j|^2}{2\,\sigma^4} \;+\; O(\rho_j^2),
  \label{eq:H_smallrho}
\end{equation}
the variance-shrinkage bracket drops to second order, only the data-driven term survives, and the noise scale $\sigma^2$ factors out of the ranking.
Define
\begin{equation}
  \colH{\tilde H_j(\theta, y)}
  \;:=\;
  \lambda_j(\theta)\;\cdot\;
  \big|\phi_j(X)^\top y\big|^{2}.
  \label{eq:Htilde}
\end{equation}
The weak-observation criterion retains both the kernel and the data factor, while $\sigma^2$ drops out: ranking depends only on the basis $\phi_j$, the kernel eigenvalues at $\theta_0$, and the data $y$.

\paragraph{A single pass over the data}
The data factor $|a_j|^2$ in $\colP{P_j}$ and $\colH{\tilde H_j}$ has a periodogram structure: it is a windowed periodogram of $y$ against $\{\phi_j\}_{j=1}^J$, and the full vector $\Phi^\top y$ is a single $\mathcal{O}(N J)$ transform of the data over the candidate basis.
In practice the user should pick $J$ as large as the candidate-precomputation budget allows: a larger $J$ widens the frequency or degree coverage of the candidate basis before any are dropped.
Since the basis $\phi_j$ never depends on $\theta$ in any of the three methods, this transform is computed once, before any hyperparameter fit, and yields $\colP{P_j}$ and $\colH{\tilde H_j}$ at any starting point $\theta_0$.
Its cost is small relative to a single hyperparameter-fit step (which itself involves an $\mathcal{O}(NM^2)$ linear solve), so it is effectively amortised by the downstream fit.

\begin{algorithm}[t]
  \caption{Selecting basis functions by score}
  \label{alg:select}
  \begin{algorithmic}[1]
    \STATE Enumerate $J \gg M$ candidate basis functions
      $\{\phi_j\}_{j=1}^J$ from the chosen basis family.
    \STATE Choose a starting point $\theta_0$ for the kernel
      hyperparameters.
    \STATE Compute the kernel weights $\lambda_j(\theta_0)$ for all
      candidates in $\mathcal{O}(J)$ time.
    \STATE Compute the data projections
      $a_j := \phi_j(X)^\top y$ for all candidates in $\mathcal{O}(N J)$ time.
    \STATE Choose one ranking score:
      \setlength{\abovedisplayskip}{5pt}\setlength{\belowdisplayskip}{5pt}%
      \setlength{\abovedisplayshortskip}{5pt}\setlength{\belowdisplayshortskip}{5pt}%
      \[
        \colI{I_j(\theta_0)} = \lambda_j(\theta_0), \qquad
        \colP{P_j(y)} = |a_j|^2, \qquad
        \colH{\tilde H_j(\theta_0, y)}
        = \lambda_j(\theta_0)|a_j|^2 .
      \]%
    \STATE Rank candidates greedily in descending order by the chosen score in $\mathcal{O}(J)$ time.
    \STATE Break ties by axis order, with axis $1$ most important.
    \STATE Return the top-$M$ indices.
  \end{algorithmic}
\end{algorithm}

\paragraph{Algorithm}
Algorithm~\ref{alg:select} summarises the selection step for any of the three criteria as a single rank-and-truncate pass.
It is important to note that the proposed criteria are heuristics.
None of them is the canonical $H_j$ at the eventual fitted hyperparameters, and the greedy rank-and-truncate approach comes with no theoretical guarantee on downstream test performance.
Whether they nonetheless do useful work is an empirical question, which the next three sections take up.

\section{Experiment I: Hilbert-space Gaussian processes}
\label{sec:hsgp}

We sweep six UCI regression benchmarks (concrete, energy, kin8nm, power, yacht, airfoil) over $M \in \{16, 32, 64, 128, 256\}$ across $10$ random $90{:}10$ train/test splits with a Mat\'ern-$5/2$ ARD kernel.
Construction, candidate-basis size, fit objective, and the non-uniform truncation baseline are in Appendix~\ref{app:hsgp-detail} (and the shared dataset and optimisation protocol in Appendix~\ref{app:expt-uci}).

\begin{figure}[t]
  \centering
  \resizebox{\linewidth}{!}{\import{figures/}{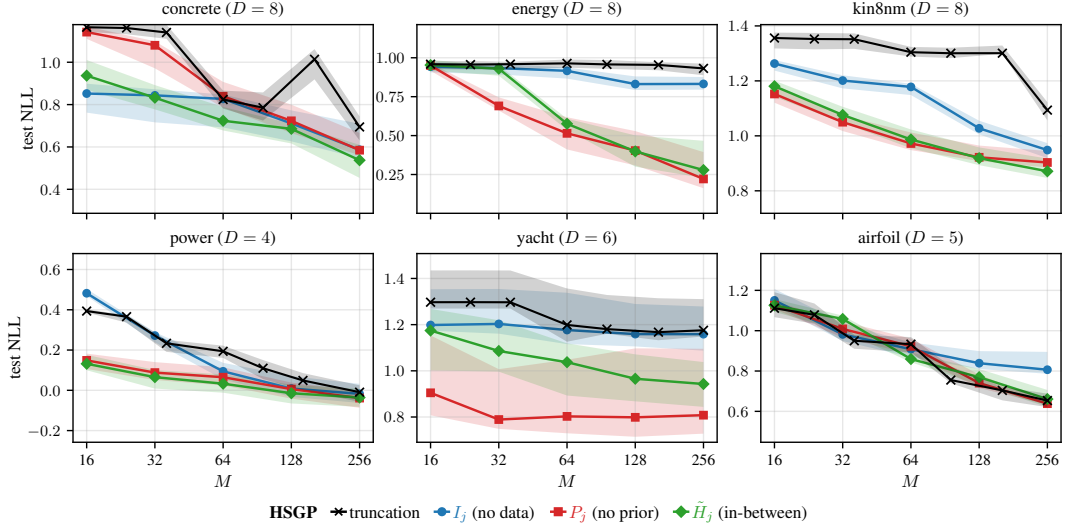}}
  \vspace{-0.8em}
  \caption{Hilbert-space Gaussian processes (HSGP) on six UCI regression benchmarks.
  Median test NLL versus compute budget $M$ across $10$ random $90{:}10$ train/test splits, with interquartile bands.
  The data-aware (in-between) $\colH{\tilde H_j}$ and (no prior) $\colP{P_j}$ improve on the (no data) $\colI{I_j}$ on most cells, with $\colP{P_j}$ slightly stronger on average.
  The non-uniform per-axis truncation baseline tracks $\colI{I_j}$ within noise, slightly worse on average.
  Truncation $x$-axis at the nearest reachable budget under $M = \prod_d M_d$ (Appendix~\ref{app:hsgp-detail}).}
  \label{fig:hsgp-uci}
\end{figure}

\paragraph{Results}
Both data-aware criteria improve on the (no data) $\colI{I_j}$ on most cells, with $\colP{P_j}$ slightly stronger on average and $\colH{\tilde H_j}$ sitting between $\colI{I_j}$ and $\colP{P_j}$ in many cells.
The improvement is largest at small $M$, where allocating budget to the basis functions the data actually activates pays off most, and shrinks toward larger $M$ as the budget catches up with the signal.
The non-uniform truncation baseline tracks $\colI{I_j}$ closely on every cell and is slightly worse on average, an empirical sanity check that the per-basis-function eigenvalue ordering is at least as good as the structural product rule.
The HSGP candidate grid is anisotropic by construction with active axes unknown before fitting, and the data projection $|a_j|^2$ gives a cheap early signal that the data-aware criteria pick up directly while $\colI{I_j}$ has to wait for the hyperparameter fit to discover it.

\section{Experiment II: Variational Fourier features}
\label{sec:vff}

We use the same six UCI benchmarks and split protocol as for HSGP, with a per-axis Mat\'ern-$5/2$ kernel, the candidate basis flattened across axes and ranked globally, and the closed-form collapsed Titsias bound fit by L-BFGS-B from the unit starting point.
Construction, the periodic-extension diagonalisation, and the per-axis truncation baseline are in Appendix~\ref{app:vff-detail}.

\begin{figure}[t]
  \centering
  \resizebox{\linewidth}{!}{\import{figures/}{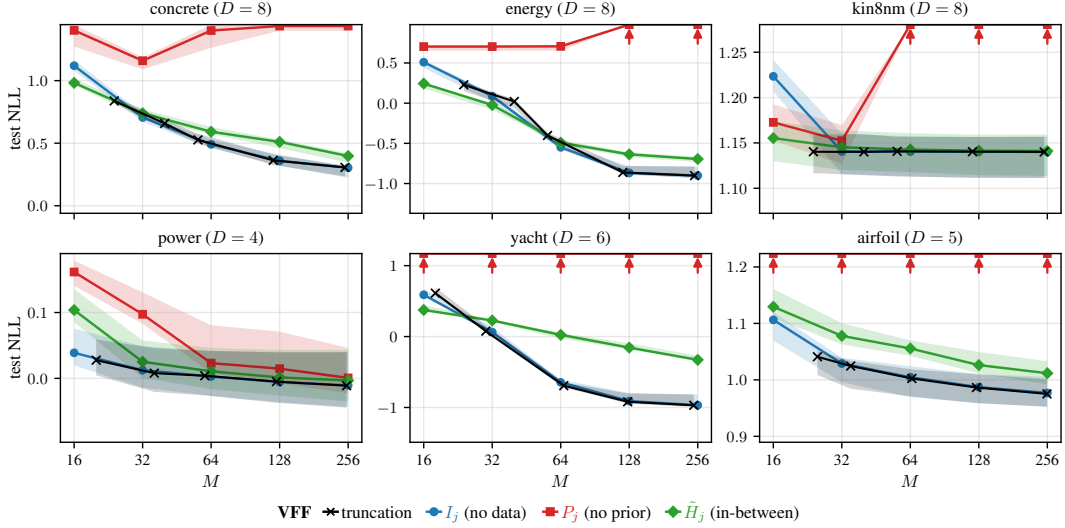}}
  \vspace{-0.8em}
  \caption{Variational Fourier features (VFF) on the same setup as Figure~\ref{fig:hsgp-uci} (Mat\'ern-$5/2$ per axis, closed-form collapsed Titsias bound).
  The data-aware (in-between) $\colH{\tilde H_j}$ and (no prior) $\colP{P_j}$ neither help nor hurt against the (no data) $\colI{I_j}$ at larger budgets, with $\colH{\tilde H_j}$ slightly behind on average.
  The conventional VFF per-axis truncation baseline sits on top of $\colI{I_j}$ at every cell.
  Truncation $x$-axis at the nearest reachable budget under $M = D(2 M_\star - 1)$ (Appendix~\ref{app:vff-detail}).}
  \label{fig:vff-uci}
\end{figure}

\paragraph{Results}
Figure~\ref{fig:vff-uci} shows that $\colH{\tilde H_j}$ is roughly flat against $\colI{I_j}$ across most cells, while $\colP{P_j}$ is reliably worse than both.
The per-axis truncation baseline sits on top of $\colI{I_j}$ at essentially every cell: the two rules pick almost identical basis-function sets, because the per-axis eigenvalue order is already the lowest-frequency-first order that truncation enforces.
On smooth UCI regressions the data projection $|a_j|^2$ mostly adds noise to an ordering that the eigenvalues already get right.

\section{Experiment III: Variational inducing spherical harmonics}
\label{sec:vish}

We use the same six UCI benchmarks and split protocol as for HSGP, with the order-$1$ arc-cosine kernel \citep{Cho2009} on the lifted inputs and the gpfy reference implementation \citep{Eleftheriadis2023} for the spherical-harmonic basis and its Funk--Hecke spectrum.
Only the basis-selection rule changes between curves, and we compare against two practitioner-default truncation rules from the literature: the cumulative-shell rule of \citet{Dutordoir2020} and the phase-truncation rule of \citet{Eleftheriadis2023}.
Construction, the closed-form arc-cosine zero pattern, and the two truncation baselines are in Appendix~\ref{app:vish-detail}.

\begin{figure}[t]
  \centering
  \resizebox{\linewidth}{!}{\import{figures/}{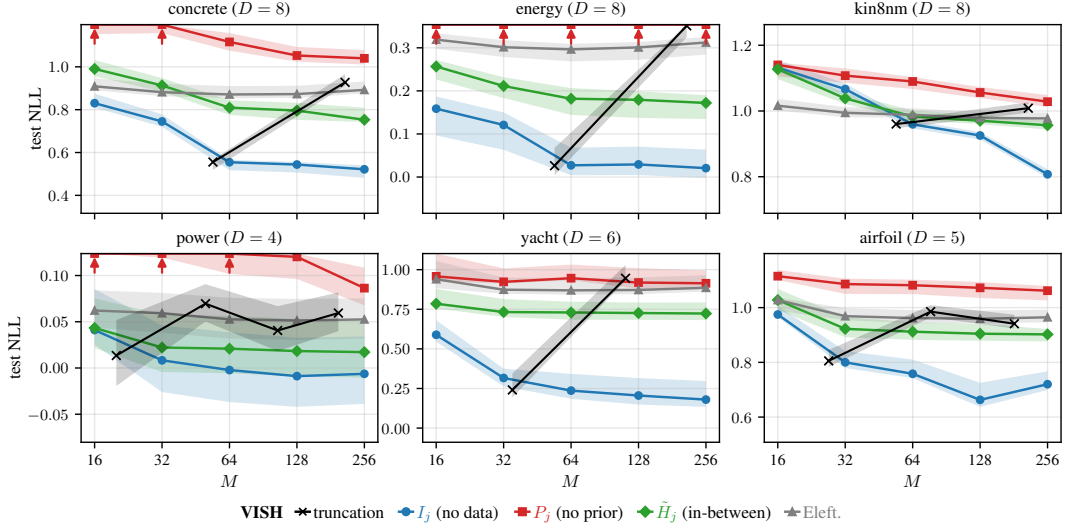}}
  \vspace{-0.8em}
  \caption{Variational inducing spherical harmonics (VISH) on the same setup as Figure~\ref{fig:hsgp-uci} (arc-cosine order-$1$ kernel, spherical harmonics on $S^{D}$).
  Against the (no data) $\colI{I_j}$, the data-aware (in-between) $\colH{\tilde H_j}$ does not improve on most cells.
  The cumulative-shell truncation curve jumps upward whenever the cutoff $L^\star$ lands on an odd degree (see body).
  Against the \citet{Eleftheriadis2023} phase-truncation baseline, every score-based criterion is consistently better, often by orders of magnitude at small $M$.
  Both truncation $x$-axes at the nearest reachable budget under each rule's structural constraint (Appendix~\ref{app:vish-detail}).}
  \label{fig:vish-uci}
\end{figure}

\paragraph{Results}
Against $\colI{I_j}$, the data-aware criteria do not help on this family.
The (no data) $\colI{I_j}$ already captures the right structural prior: spherical-harmonic eigenvalues drop fast with degree, and on smooth UCI regressions the signal energy is concentrated at low degree, so reordering by data energy mostly perturbs an already-good ranking.
The cumulative-shell truncation curve, however, behaves more dramatically: at cumulative-shell budgets that end on an even degree it selects bit-for-bit the same basis functions as $\colI{I_j}$, but whenever the cutoff $L^\star$ lands on an odd degree $\ell \ge 3$ the curve jumps upward.
The cause is a closed-form property of the order-$1$ arc-cosine kernel: its Funk--Hecke expansion has $\mu_\ell = 0$ for all odd $\ell \ge 3$ \citep[Appendix~D.2]{Bach2017}, independent of the sphere dimension $D$, so filling an odd shell spends budget on coefficients the kernel forces to zero variance, and the (no data) ranking skips those shells automatically.
Against the phase-truncation baseline, every score-based criterion is consistently better, with the gap reaching one to two orders of magnitude in test NLL at small $M$ on energy and power: the single global $m^\star$ knob cannot adapt when the data favour more basis functions at one degree than another, and it inherits the same odd-shell trap.
The phase-truncation sweep also takes about $10.5$ seconds per fit versus $1.4$ seconds for the score-selected VISH sweep, with about $18\times$ as many L-BFGS-B iterations, because the phase variables are optimised inside the fit whereas $\colI{I_j}$ fixes a subset before fitting.

\section{Related work}
\label{sec:related}

\paragraph{Active-set and information-theoretic selection in Gaussian processes}
Sparse-GP active-set methods have long used greedy and information-theoretic criteria to decide which scalar linear functionals of $f$ enter the approximation.
The informative vector machine of \citet{lawrence2002fast}, fast forward selection for sparse GP regression \citep{Seeger2003}, sparse online GPs \citep{csato2002sparse}, and sparse greedy GP regression \citep{smola2001sparse} all construct a small active set this way, and the same KL information gain $D_{\mathrm{KL}}(p(u_j \mid y) \,\|\, p(u_j))$ that we use for $H_j$ underlies several of these scores \citep{rasmussen2006gaussian, krause2008near}.
More recent work continues this line by placing Bayesian or task-specific structure on inducing locations and their number, for example through point-process priors \citep{uhrenholt2021probabilistic}, Bayesian inference over inducing locations \citep{Rossi2021}, or quality-diversity allocation for Bayesian optimisation \citep{moss2023inducing}.
For continuous inducing-point locations $z \in \mathcal{X}$ the criterion is differentiable and is typically optimised jointly with the variational posterior \citep{snelson2006sparse}, but all of these methods select, infer, or allocate data-indexed inducing inputs as a sub-routine of inference itself.
Our setting is different.
The basis families we consider already define a fixed candidate set of linear functionals (HSGP eigenfunctions, VFF Fourier features, VISH spherical harmonics), and the question is only how to spend a given budget $M$ inside that set.
Once this candidate basis is fixed, each criterion is a scalar function of the prior weight $\lambda_j$ and the data projection $a_j = \phi_j(X)^\top y$, and all candidates can be ranked in one pass before any posterior fit.
The contribution is therefore not a new active-set sparse-GP algorithm, but a basis-index selection rule that can replace the default truncation rule inside existing basis-function sparse-GP constructions.

\paragraph{Matching pursuit and correlation screening}
The (no prior) score $\colP{P_j} = |a_j|^2$ is the same correlation-screening step that initialises matching pursuit and its orthogonal variant \citep{mallat1993matching, pati1993orthogonal}, in which dictionary atoms are picked by their alignment with the signal or current residual.
Two differences matter.
Matching pursuit is iterative: after an atom is selected the residual is updated, and in orthogonal matching pursuit the selected atoms are re-orthogonalised.
Our selection step is a non-iterative rank-and-truncate pass that fixes the GP basis once, before any fit.
The GP setting also supplies a prior variance $\lambda_j$ for each candidate coefficient, so the (no data) $\colI{I_j} = \lambda_j$ and (in-between) $\colH{\tilde H_j} = \lambda_j |a_j|^2$ are not pure correlation-screening rules.
They combine the observed alignment with the kernel-implied plausibility of the candidate, which is what makes the same selection idea meaningful across HSGP, VFF, and VISH, whose basis indices have different geometries but all carry GP prior weights.

\paragraph{Spectral and eigenfunction feature methods}
A different line of work changes the feature construction itself.
Sparse-spectrum GPs choose a small set of spectral frequencies and learn their locations as model parameters \citep{Lazaro-Gredilla2010}.
Variational orthogonal features construct stationary-kernel features for cheaper variational inference \citep{Burt2020}, and data-dependent eigenfunction methods select eigenfunctions of an empirical Gram matrix by evidence maximisation \citep{qi2010eigengp}.
These ask which feature family or spectral approximation should define the approximation.
Our setting is narrower but complementary: the feature family is fixed by a published sparse-GP construction, and we only choose the subset of its candidate indices.

\section{Discussion}
\label{sec:disc}

The work done here is essentially a preliminary investigation of a gap in the sparse-GP literature.
The three methods we considered all recommend truncation as the default way to spend the basis budget.
Yet for a fixed compute budget $M$ (the realistic inference setting), there is freedom in choosing \emph{which} basis functions to actually use, and the truncation default does not exercise it.
Asking whether the candidate basis should be selected instead turns out to be cheap enough to test, and informative enough to separate from a broad benchmark of sparse-GP approximations.
The investigation has taught us two things.

The first is that the (no data) criterion $\colI{I_j}$ is a strong default: ranking every candidate freely by its kernel-prior weight is at least as good as the standard truncation rule for each method.
On VISH it is indeed much better than either of the two practitioner-default truncation rules we compared against:
an odd-shell zero pattern in the order-$1$ arc-cosine kernel kills entire structural blocks that the standard truncation rules waste budget on, while $\colI{I_j}$ avoids them automatically (Section~\ref{sec:vish}).
The same simple ranking also defeats the phase-truncation baseline of \citet{Eleftheriadis2023}, with substantial speedups in addition to the accuracy gain (Section~\ref{sec:vish}).

\begin{figure}[t]
  \centering
  \resizebox{\linewidth}{!}{\import{figures/motivational/}{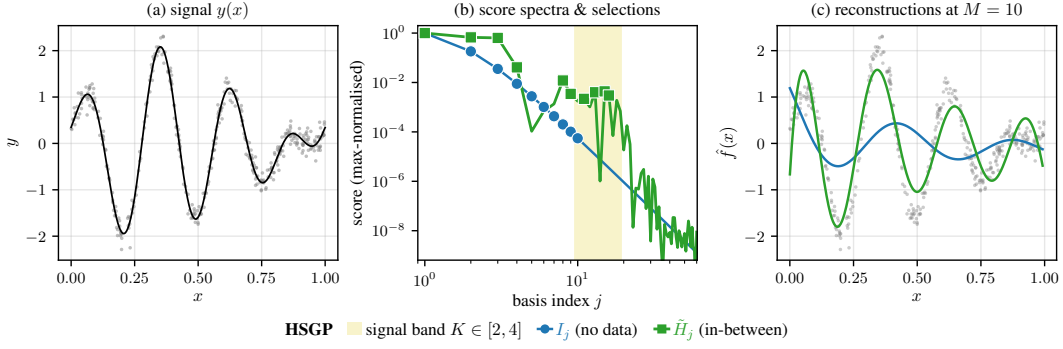}}
  \vspace{-0.8em}
  \caption{Toy bandpass example (details in Appendix~\ref{app:toy}).
  (a) Signal in the integer-frequency band $K \in [2, 4]$ plus noise.
  (b) Score spectra: the (no data) $\colI{I_j}$ is monotone low-pass and picks the lowest frequencies (circles), while the (in-between) $\colH{\tilde H_j}$ rises inside the band and picks there (squares).
  (c) HSGP fits at $M = 10$: $\colI{I_j}$ over-smooths through the band, while $\colH{\tilde H_j}$ tracks the wiggles.}
  \label{fig:bandpass}
\end{figure}

The second is that the data-aware (no prior) and (in-between) criteria are method-specific.
On HSGP the candidate grid (a high-dimensional version of the rectangle in Figure~\ref{fig:motivational}) has unknown active axes before fitting, so the data projection $|a_j|^2$ is the cheapest probe of where the signal lives, and both $\colP{P_j}$ and $\colH{\tilde H_j}$ give an essentially free improvement over truncation.
On VFF and VISH the basis is naturally ordered to favour low-frequency or low-degree functions, which is what smooth UCI regressions need, and $\colI{I_j}$ alone exploits this.
The data-aware criteria are most likely to help when the data has structure that low-frequency truncation misses, as in the toy bandpass example of Figure~\ref{fig:bandpass}.

\paragraph{Limitations and future work}
The eigenvalue-dependent criteria $\colI{I_j}$ and $\colH{\tilde H_j}$ evaluate $\lambda_j$ at a single guessed starting point $\theta_0$ rather than integrating over plausible hyperparameters, and the closed-form $H_j$ is exact only under the empirical-decoupling assumption $\Phi^\top \Phi \approx \mathrm{diag}(\|\phi_j(X)\|^2)$.
The three methods also differ in how their basis is constructed (HSGP native multi-dimensional, VFF additive across axes, VISH native multi-dimensional on the lifted sphere), so cross-method differences in selection behaviour can reflect this construction confound as well as the criteria.
The empirics are preliminary, with six UCI regressions at five budgets per method and test RMSE corroborating the test-NLL pattern (Appendix~\ref{app:rmse-scores}).
Whether the patterns generalise to other kernels, basis families, or larger-scale regressions remains for future work.

\paragraph{Conclusion}
This paper addressed a gap in the basis-function sparse-GP literature: the practitioner has a fixed compute budget $M$ to spend across the candidate basis, but the default answer is structural truncation.
We asked whether a simple one-pass selection criterion can do better, and the answer turns out to be yes, with the size of the win depending on the method.
The (no data) $\colI{I_j}$ ties or beats truncation everywhere and substantially improves over it on VISH, while the data-aware (no prior) $\colP{P_j}$ and (in-between) $\colH{\tilde H_j}$ criteria further improve on HSGP, which is the most adopted method in practice.

\bibliographystyle{plainnat}
\bibliography{main}

\appendix
\counterwithin{figure}{section}
\setcounter{figure}{0}
\renewcommand{\theHfigure}{\Alph{section}.\arabic{figure}}

\section{Sparse GPs and basis-function expansions}
\label{app:setup}

The main text uses the language of finite basis expansions because that is where the selection problem is most transparent: once a candidate dictionary has been fixed, one has to decide which coefficients to keep.
Sparse GPs in the sense of \citet{Leibfried2022}, however, are slightly broader objects than finite expansions.
We spell out the connection here so that the basis-expansion view used throughout the paper is not mistaken for the most general sparse-GP construction.

\paragraph{A GP is already a basis-function expansion}
A Gaussian process $f \sim \mathcal{GP}(0, k)$ on a domain $\mathcal{X}$ admits a Mercer expansion of its kernel,
\[
  k(x, x') \;=\; \sum_{j=1}^{\infty} \lambda_j\, \phi_j(x)\, \phi_j(x'),
\]
in an orthonormal basis $\{\phi_j\}_{j=1}^{\infty}$ of $L^2(\mathcal{X}, \mu)$ for a natural measure $\mu$, with non-negative eigenvalues $\lambda_1 \ge \lambda_2 \ge \cdots$.
By Karhunen--Lo\`eve, a draw $f \sim \mathcal{GP}(0, k)$ is then
\[
  f(x) \;=\; \sum_{j=1}^{\infty} w_j\, \phi_j(x), \qquad w_j \overset{\mathrm{iid}}{\sim} \mathcal{N}(0, \lambda_j).
\]
Thus a GP is already a basis-function expansion.
What is infinite-dimensional is the number of random coefficients, not the expansion form itself.

\paragraph{The general sparse-GP construction}
The general definition of \citet{Leibfried2022} does not require Mercer.
A sparse GP starts from a GP $f \sim \mathcal{GP}(0, k)$ conditioned on a finite set of inducing variables $\mathbf{u} = (u_1, \ldots, u_M)$, each a scalar linear functional of $f$, together with some chosen distribution $q(\mathbf{u}) = \mathcal{N}(\mathbf{m}_u, S_{uu})$.
Marginalising out gives the sparse-GP marginal
\[
  f(\cdot) \;\sim\; \mathcal{GP}\!\Big(\mu(\cdot) + k_{\cdot u}\, K_{uu}^{-1}(\mathbf{m}_u - \mu_u),\; k(\cdot, \cdot') - k_{\cdot u}\, K_{uu}^{-1}\,(K_{uu} - S_{uu})\, K_{uu}^{-1}\, k_{u\cdot'}\Big).
\]
The covariance is the prior kernel minus a rank-$M$ correction, not a rank-$M$ kernel itself, and so the random function $f$ retains infinite-dimensional uncertainty around the conditional mean.
A sparse GP per Leibfried is therefore not a finite basis-function expansion in general.

\paragraph{Mercer connects the two}
Take the inducing variables to be Mercer coefficients, $u_j := \langle f, \phi_j \rangle$, for the first $M$ Mercer eigenfunctions.
By orthonormality and Mercer, $K_{uu} = \mathrm{diag}(\lambda_1, \ldots, \lambda_M)$ and $k_{xu}[j] = \lambda_j \phi_j(x)$.
Substituting into Leibfried's covariance with $S_{uu} = K_{uu}$, the sparse-GP covariance becomes the tail of the Mercer expansion,
\[
  \mathrm{Cov}_{\text{sparse}}(f(x), f(x')) \;=\; \sum_{j > M} \lambda_j\, \phi_j(x)\, \phi_j(x'),
\]
and the GP itself decomposes as
\[
  f(x) \;=\; \underbrace{\sum_{j=1}^{M} u_j\, \phi_j(x)}_{\text{rank-}M \text{ basis expansion}} \;+\; \underbrace{h(x)}_{\text{Mercer tail}},
\]
with $u_j \sim \mathcal{N}(0, \lambda_j)$ and $h$ a zero-mean GP whose covariance is the tail of the Mercer expansion.
The finite basis-function expansion view drops the residual $h$, whereas the full sparse-GP view keeps it.
Both are valid sparse-GP constructions; they differ only on the orthogonal complement of $\mathrm{span}\{\phi_j\}_{j=1}^{M}$, on which the basis expansion sets $f$ to zero while the sparse GP keeps the full prior.

\paragraph{What each family in this paper does}
HSGP \citep{Solin2020, Riutort-Mayol2020} drops the residual: the model is an explicit basis-function expansion, with Dirichlet Laplacian eigenfunctions $\phi_j$ and prior variances $\lambda_j(\theta)$ from the kernel's spectral density, and inference is closed-form Bayesian linear regression on the truncated basis.
VFF \citep{Hensman2017} keeps the residual: the inducing variables are RKHS Fourier projections, and Hensman et al.\ comment explicitly that the variational form does not discard the orthogonal complement.
VISH \citep{Dutordoir2020, Eleftheriadis2023} keeps the residual on the same logic with spherical-harmonic projections.
Appendices~\ref{app:hsgp-detail}--\ref{app:vish-detail} give the construction of each family.

\paragraph{Why the residual does not affect our criteria}
The criteria $\colI{I_j(\theta)}$, $\colP{P_j(y)}$, $\colH{\tilde H_j(\theta, y)}$ are derived from the per-basis-function KL information gain $H_j(\theta, y) = D_{\mathrm{KL}}\!\big(p(u_j \mid y) \,\|\, p(u_j)\big)$.
The prior $p(u_j) = \mathcal{N}(0, \lambda_j(\theta))$ depends only on the kernel eigenvalue at $\phi_j$, not on whether the residual is kept or dropped.
The marginal posterior $p(u_j \mid y)$ depends on the data through the projection $a_j = \phi_j(X)^\top y$ and on the noise scale $\sigma^2$, and under the empirical-decoupling assumption $\Phi^\top \Phi \approx \mathrm{diag}(c_j)$ takes the same value in both pictures: the residual $h$ contributes only to the unmodelled uncertainty in $f(x)$ at unseen test inputs, not to the marginal posterior on the $j$-th Mercer coefficient.
The criteria, and the selection problem they pose, are therefore identical in both pictures.

\paragraph{Conventions used in the main text}
Given this equivalence at the criterion level, the main text uses the basis-expansion picture as its default exposition.
It poses the selection-versus-truncation question cleanly (which Mercer coefficients to keep), unifies the three families at the model-and-prior level (each picks $\{\phi_j\}$ and $\lambda_j(\theta)$), and does not commit to any inference machinery.
The full sparse-GP construction is invoked per family only when reporting the fitting procedure each family uses downstream of selection: the closed-form marginal likelihood for HSGP, and the collapsed Titsias bound for VFF and VISH.

\section{Derivation of the criteria}
\label{app:derivation}

We derive the closed form Equation~\eqref{eq:H_full}, the data-averaged form $\mathbb{E}_y[H_j] = \tfrac{1}{2}\log(1 + \rho_j)$, and the rank-equivalence of the fully data-fit $H_j$ with the no-prior criterion $\colP{P_j}$.
The same argument extends to arbitrary scalar linear functionals of $f$ (inducing points, Mercer projections, windowed Fourier projections).
The basis-function specialisation used in the main text follows.

\paragraph{Closed form for $H_j$}
Under the empirical-decoupling assumption $\Phi^\top \Phi \approx \mathrm{diag}(c_j)$ with $c_j := \|\phi_j(X)\|^2$, the marginal posterior on the basis-function coefficient $u_j$ is Gaussian with variance $v_j^{\text{post}} = v_j / (1 + \rho_j)$ and mean $\mu_j(y) = \lambda_j a_j / (\sigma^2 + \lambda_j c_j)$, where $v_j = \lambda_j(\theta)$ is the prior variance and $\rho_j = \lambda_j c_j / \sigma^2$ is the per-basis-function signal-to-noise ratio.
Plugging into the univariate Gaussian KL gives Equation~\eqref{eq:H_full}.

\paragraph{Data-averaged form: derivation of $I_j$}
Under the prior, $a_j = \phi_j(X)^\top y$ has variance $\mathrm{Var}(a_j) = \sigma^2 c_j (1 + \rho_j)$, so $\mathbb{E}_y[|a_j|^2] = \sigma^2 c_j (1 + \rho_j)$.
Substituting into the data-driven term in Equation~\eqref{eq:H_full},
\[
  \mathbb{E}_y\!\left[\frac{\lambda_j |a_j|^2}{2 \sigma^4 (1+\rho_j)^2}\right]
  =
  \frac{\lambda_j c_j}{2 \sigma^2 (1 + \rho_j)}
  =
  \frac{\rho_j}{2(1+\rho_j)},
\]
and adding the variance-shrinkage bracket $\tfrac{1}{2}[\log(1+\rho_j) - \rho_j/(1+\rho_j)]$ collapses the data-driven and shrinkage pieces:
\[
  \mathbb{E}_y[H_j(\theta, y)]
  \;=\;
  \tfrac{1}{2}\log(1 + \rho_j),
\]
which is monotone in $\rho_j$ at fixed $\sigma^2, c_j$ and therefore equivalent in rank to $\lambda_j(\theta)$.
This is the criterion $\colI{I_j}$ of Equation~\eqref{eq:I}.

\paragraph{Rank-equivalence of the fully data-fit $H_j$ with $\colP{P_j}$}
If $\sigma^2$ and $\lambda_j$ are jointly fit by maximum likelihood at fixed $j$, the optimum gives $\hat\rho_j = \max(0, |a_j|^2 / (\sigma^2 c_j) - 1)$.
Substituting back into Equation~\eqref{eq:H_full} reduces $H_j$ to a monotone function of $|a_j|^2$ alone, which is rank-equivalent to $\colP{P_j} = |a_j|^2$.

\section{Hilbert-space Gaussian processes (HSGP)}
\label{app:hsgp-detail}

For HSGP \citep{Solin2020, Riutort-Mayol2020}, the candidate dictionary is fixed by the geometry of a box and a Dirichlet boundary condition, while the kernel enters through spectral weights on that dictionary.
The details below give the basis $\{\phi_j\}$, the prior variances $\lambda_j(\theta)$, the downstream fit after selection, and the per-axis truncation baseline reported in Section~\ref{sec:hsgp}.

\paragraph{The basis on a box}
Take the input domain to be $\Omega = [-L, L]^D$.
The Dirichlet eigenproblem for the Laplacian on $\Omega$,
\[
  -\nabla^2 \phi_j(x) = \mu_j\, \phi_j(x), \qquad x \in \Omega, \qquad \phi_j\big|_{\partial \Omega} = 0,
\]
has eigenfunctions
\[
  \phi_j(x) = \prod_{d=1}^{D} L^{-1/2}\, \sin\!\left(\frac{\pi j_d (x_d + L)}{2L}\right), \qquad j = (j_1, \ldots, j_D) \in \mathbb{N}_+^{D},
\]
with eigenvalues
\[
  \mu_j = \sum_{d=1}^{D} \!\left(\frac{\pi j_d}{2L}\right)^2.
\]
The $\{\phi_j\}$ are orthonormal in $L^2(\Omega)$ and depend only on $\Omega$ and the boundary condition.
In particular they do not depend on the kernel hyperparameters $\theta$.

\paragraph{Approximate Mercer interpretation}
For a stationary kernel $k_\theta$ on $\mathbb{R}^D$ with spectral density $S_\theta$, the HSGP construction approximates the kernel by
\[
  k_\theta(x, x') \;\approx\; \sum_j \lambda_j(\theta)\, \phi_j(x)\, \phi_j(x'), \qquad \lambda_j(\theta) = S_\theta(\omega_j),
\]
where $\omega_j = (\pi j_1 / 2L_1, \ldots, \pi j_D / 2L_D)$ is the per-axis Laplacian frequency vector (the box $\Omega = \prod_d [-L_d, L_d]$ is allowed to be anisotropic, and the kernel can be ARD).
The approximation comes from restricting to a compact $\Omega$, imposing Dirichlet boundary conditions, and truncating to finitely many modes.
It is asymptotically exact as the per-axis box lengths $L_d \to \infty$ for inputs away from $\partial \Omega$ and with sufficient mode count.
Throughout this paper we use the box $L_d = 1.2 \cdot \max_i |x_{id}|$ per axis, following the convention of \citet{Riutort-Mayol2020}.

\paragraph{Inference under Gaussian likelihood}
Once a subset $\mathcal{M}$ of size $M$ is selected, the model is the finite linear-Gaussian system $y = \Phi u + \varepsilon$ with $\Phi_{nj} = \phi_j(x_n)$, basis-function coefficients $u = (u_j)_{j \in \mathcal{M}} \sim \mathcal{N}(0, \Lambda_\theta)$, $\Lambda_\theta = \mathrm{diag}(\lambda_j(\theta))_{j \in \mathcal{M}}$, and $\varepsilon \sim \mathcal{N}(0, \sigma^2 I_N)$.
The marginal likelihood,
\[
  \log p(y \mid \theta, \mathcal{M}) = -\tfrac{1}{2}\!\left[\log|\Sigma_y| + y^\top \Sigma_y^{-1} y + N\log 2\pi\right], \qquad \Sigma_y = \Phi \Lambda_\theta \Phi^\top + \sigma^2 I_N,
\]
is closed-form, so the downstream fit is L-BFGS-B maximisation of this objective rather than variational inference.

\paragraph{Truncation baseline}
The conventional HSGP library default fixes per-axis counts $(M_1, \ldots, M_D)$ and keeps every basis function with $j_d \le M_d$ on every axis, giving a total of $M = \prod_d M_d$ basis functions.
For our budget range $M \in [16, 256]$ this rule is coarse-grained: a single uniform $M_d = M_\star$ on every axis only lands on $M = M_\star^D$, which covers at most one or two budgets per dataset.
We therefore use a non-uniform extension to obtain a fair comparison curve on the same $M$-grid as the selection criteria.
For each target budget we pick the most-uniform integer factorisation $\prod_d M_d$, breaking ties by smallest standard deviation across axes, and assign the descending-sorted tuple to axes ordered by descending training-input variance.
This defines a reproducible non-uniform truncation baseline on the same $M$-grid as the selection criteria, used in Figure~\ref{fig:hsgp-uci}.

\section{Variational Fourier features (VFF)}
\label{app:vff-detail}

For VFF \citep{Hensman2017}, the candidate basis is Fourier rather than Laplacian, and the multi-dimensional candidate basis used here is assembled from separate one-dimensional Fourier blocks rather than from their tensor product.
We describe the one-dimensional features, the diagonalisation approximation used for scoring, the Gaussian-likelihood objective, and the per-axis truncation baseline used in Section~\ref{sec:vff}.

\paragraph{Fourier features on an interval}
Take the input domain to be $[a, b]$ with length $T = b - a$, and use the normalised measure $d\mu(x) = dx/T$.
Define harmonic frequencies $\omega_m = 2\pi m / T$ for $m = 1, 2, \ldots$ and the orthonormal Fourier features
\[
  \phi_0(x) = 1, \qquad \phi_{c, m}(x) = \sqrt{2}\cos\!\big(\omega_m (x - a)\big), \qquad \phi_{s, m}(x) = \sqrt{2}\sin\!\big(\omega_m (x - a)\big).
\]
These are orthonormal in $L^2([a, b], \mu)$ and depend only on the interval and the frequency grid.
For the multi-dimensional case we use $f(x) = \sum_{d=1}^{D} f_d(x_d)$ with a separate 1D VFF basis per axis.
The total candidate count is $\sum_d M_d$, not the tensor-product $\prod_d M_d$, so the construction scales linearly in $D$ at fixed per-axis count.

\paragraph{Periodic-extension diagonalisation}
The Fourier features are not exact Mercer eigenfunctions of a stationary kernel on a finite interval: boundary effects prevent perfect diagonalisation.
\citet{Hensman2017} show that for Mat\'ern kernels the exact $K_{uu}$ is diagonal-plus-low-rank, with the diagonal proportional to $1/S_\theta(\omega_m)$ and rank-one corrections capturing the boundary effects.
We adopt the standard periodic-extension approximation, in which the kernel is extended periodically with period $T$ and diagonalises in the Fourier basis with spectral weights
\[
  \lambda_m(\theta) \;\approx\; \frac{S_\theta(\omega_m)}{T}.
\]
The approximation is asymptotically exact as $T$ grows large relative to the kernel's correlation length, and inherits boundary error at finite $T$.
For our criteria the diagonal piece suffices because the criteria depend only on the per-basis-function eigenvalue $\lambda_m$ and the data projection.

\paragraph{Inference under Gaussian likelihood}
A Gaussian variational posterior $q(\mathbf{u}) = \mathcal{N}(\mathbf{m}, S)$ on the inducing variables $\mathbf{u} = (u_j)_{j \in \mathcal{M}}$, the per-feature coefficients with prior variance $\lambda_j(\theta)$, is fit by maximising the standard sparse-GP ELBO.
For Gaussian likelihood the optimum admits a closed form (the collapsed Titsias bound), and the bound itself becomes the marginal-likelihood objective that we maximise over $\theta$ by L-BFGS-B \citep{Titsias2009, Hensman2017}.
The closed-form gradients and predictive expressions follow Hensman et al.\ directly.

\paragraph{Truncation baseline}
The conventional VFF library default fixes a per-axis frequency count $M_\star$ and keeps frequencies $j_d \in \{0, 1, \ldots, M_\star - 1\}$ on every axis.
Each non-DC frequency contributes both a cosine and a sine basis function while $j_d = 0$ contributes only a cosine, so the total count is $M = D(2 M_\star - 1)$.
For each dataset we sweep $M_\star$ such that $M \in [16, 256]$ and thin to five log-spaced points (e.g.\ on \emph{combined cycle power plant} with $D = 4$ this lands at $M \in \{20, 36, 60, 124, 252\}$, and on \emph{kin8nm} with $D = 8$ at $M \in \{24, 40, 56, 120, 248\}$).
This is the truncation curve reported in Figure~\ref{fig:vff-uci}.

\section{Variational inducing spherical harmonics (VISH)}
\label{app:vish-detail}

For VISH \citep{Dutordoir2020, Eleftheriadis2023}, the candidate dictionary is organised by spherical-harmonic degree after the Euclidean inputs have been lifted to the sphere.
The details below give that construction, the order-$1$ arc-cosine kernel \citep{Cho2009} used throughout, the closed-form zero pattern responsible for the cumulative-shell jump in Figure~\ref{fig:vish-uci}, and the two truncation baselines used in Section~\ref{sec:vish}.

\paragraph{Input lifting and spherical harmonics}
Given Euclidean inputs $x_n \in \mathbb{R}^D$, define
\[
  z_n \;=\; \frac{(x_n, 1)}{\|(x_n, 1)\|} \;\in\; S^{D} \;\subset\; \mathbb{R}^{\bar D},
  \qquad \bar D \;:=\; D + 1.
\]
All harmonic features are evaluated at $z_n$, not at the raw Euclidean input.
The lift raises the dimension by one, so the sphere $S^{D}$ has ambient dimension $\bar D$, and it is $\bar D$, not $D$, that enters the harmonic-analysis formulas below.
Spherical harmonics $\phi_{\ell, k} : S^{D} \to \mathbb{R}$ are indexed by degree $\ell = 0, 1, 2, \ldots$ and orientation $k = 1, \ldots, N(\bar D, \ell)$, with multiplicity
\[
  N(\bar D, \ell) \;=\; \frac{(2\ell + \bar D - 2)\,(\ell + \bar D - 3)!}{\ell!\,(\bar D - 2)!}, \qquad \bar D \ge 3,
\]
and are orthonormal under the normalised surface measure $d\mu(z) = d\omega(z)/\Omega_{D}$, with $\Omega_{D}$ the surface area of $S^{D}$.

\paragraph{Funk--Hecke and degree-only eigenvalues}
A kernel $k_\theta$ on $S^{D}$ is zonal if $k_\theta(z, z') = \kappa_\theta(z^\top z')$, i.e.\ depends only on the angle between inputs.
For a zonal kernel, the Funk--Hecke theorem says that spherical harmonics are eigenfunctions of the kernel integral operator with eigenvalues that depend only on the degree:
\[
  \lambda_{\ell, k}(\theta) \;=\; \lambda_\ell(\theta) \quad\text{for all } k = 1, \ldots, N(\bar D, \ell).
\]
This gives a shell decomposition indexed by degree $\ell$, where all $N(\bar D, \ell)$ orientations within a shell share the same prior variance.

\paragraph{The order-1 arc-cosine kernel}
We use the order-$1$ arc-cosine kernel \citep{Cho2009} throughout, with $\lambda_\ell(\theta)$ obtained from the gpfy reference implementation \citep{Eleftheriadis2023}.
A closed-form property of this kernel is that its Funk--Hecke expansion has $\lambda_\ell = 0$ for all odd $\ell \ge 3$, independent of the ambient dimension $\bar D$ \citep[Appendix~D.2]{Bach2017}.
Only the shells $\ell \in \{0, 1\} \cup \{2, 4, 6, \ldots\}$ carry positive prior variance.
In the dimensions used by the lifted UCI datasets ($\bar D \in \{5, 6, 7, 9\}$), the same pattern is visible numerically: the magnitudes of the nonzero shells decay smoothly with $\ell$, while the odd shells $\ell \ge 3$ sit at machine precision.
The pattern is a property of the kernel function, not the sphere geometry, and it has a sharp practical consequence for the cumulative-shell truncation baseline below.

\paragraph{Inference under Gaussian likelihood}
Inducing variables are RKHS projections, $u_{\ell, k} = \langle f, \phi_{\ell, k}\rangle_{\mathcal{H}}$.
The prior $K_{uu}$ is diagonal in the spherical-harmonic basis with entries $1/\lambda_\ell$ in the RKHS coordinate.
The Mercer coordinate $c_{\ell, k} = \lambda_\ell u_{\ell, k}$ has prior variance $\lambda_\ell$, which is what the criteria use.
A Gaussian variational posterior $q(\mathbf{u}) = \mathcal{N}(\mathbf{m}, S)$ is fit by maximising the standard sparse-GP ELBO, which for Gaussian likelihood admits a closed form.
We maximise the resulting collapsed bound over $\theta$ by L-BFGS-B \citep{Dutordoir2020, Titsias2009}.
The diagonal $K_{uu}$ structure is preserved under arbitrary subset selection.

\paragraph{Two truncation baselines}
The VISH comparisons use the two practitioner-default truncation rules from the literature.

The cumulative-shell rule of \citet{Dutordoir2020} fills harmonic shells in degree order: pick a cutoff degree $L^\star$ and keep every harmonic at $\ell \le L^\star$, for a total of $M = \sum_{\ell = 0}^{L^\star} N(\bar D, \ell)$ basis functions.
This is the direct VISH analogue of the per-axis cube rule used for HSGP.
On the order-$1$ arc-cosine kernel, however, this rule can select zero-variance shells: whenever $L^\star$ is an odd degree $\ell \ge 3$, every harmonic at that degree has zero prior variance and contributes no information to the fit.
This is the cause of the upward jump in the black truncation curve in Figure~\ref{fig:vish-uci}.

The phase-truncation rule of \citet{Eleftheriadis2023} keeps $\min(m^\star, N(\bar D, \ell))$ orthogonalised phase features per shell up to a larger cutoff degree, where the phases are parameterised by learnable phase vectors and computed via the addition theorem as Gegenbauer evaluations $C_\ell^{(\alpha)}(z^\top v_{\ell, m})$ with $\alpha = (\bar D - 2)/2$.
This rule has a single integer knob $m^\star$ on top of the cutoff, and we choose $m^\star$ so that the resulting basis-function count is closest to the target $M$.
It is data-blind (the phases are optimised at fitting time, not at selection time) and inherits the same odd-shell trap as cumulative-shell truncation.

\section{Toy figure details}
\label{app:toy}

The two illustrative figures in the main text serve a narrower purpose than the UCI sweeps: they make visible what a ranking rule is doing before any benchmark comparison is considered.
We give the data-generation and fitting details for Figure~\ref{fig:motivational} (the four-panel grid in Section~\ref{sec:intro}) and Figure~\ref{fig:bandpass} (the one-dimensional bandpass example in Section~\ref{sec:score}) so that the examples can be read as concrete constructions rather than as additional empirical claims.

\paragraph{Figure~\ref{fig:motivational} (motivational grid, two-dimensional HSGP)}
We use the AT and V features (axes $0$ and $1$) of the UCI \emph{combined cycle power plant} regression \citep{ucimlr} to build a $D = 2$ HSGP problem with $M_{\text{per-dim}} = 5$, giving a $5 \times 5 = 25$ candidate basis on a box $[-L_d, L_d]^2$ with $L_d = 1.2 \cdot \max_i |x_{id}|$ per axis.
The kernel is Mat\'ern-$5/2$ with ARD lengthscales.
For each of the four selection strategies (rectangular truncation $M_d = (5, 2)$, the (no data) criterion $\colI{I_j}$, the (no prior) criterion $\colP{P_j}$, the (in-between) criterion $\colH{\tilde H_j}$), we select the top-$M = 10$ candidates at the unit starting point $\theta_0$ and fit the kernel hyperparameters by L-BFGS-B on the closed-form marginal likelihood from the same starting point, reporting test NLL on the held-out split.

\paragraph{Figure~\ref{fig:bandpass} (one-dimensional bandpass example)}
We sample $N$ noisy observations of a synthetic bandpass-sinc signal on $t \in [0, T]$ at sampling frequency $f_s = 512$ ($T = 1$, $N = 512$ regularly-spaced points), with bandpass cutoffs $\omega_0 = 30$ and $\omega_1 = 100$ and additive Gaussian noise of standard deviation $\sigma_{\text{noise}} = 0.3$.
The HSGP basis uses $M_{\text{cand}} = 512$ Dirichlet sine eigenfunctions on a box of length $1.2 \cdot T$.
The score-spectrum panels show $\colI{I_j}$ and $\colH{\tilde H_j}$ at $M = 10$ alongside the corresponding HSGP fits.

\section{Experimental details for UCI}
\label{app:expt-uci}

\paragraph{Datasets and splits}
Six UCI regression benchmarks \citep{ucimlr}: airfoil self-noise ($N = 1503$, $D = 5$), concrete compressive strength ($N = 1030$, $D = 8$), energy efficiency ($N = 768$, $D = 8$), kin8nm ($N = 8192$, $D = 8$), combined cycle power plant ($N = 9568$, $D = 4$), and yacht hydrodynamics ($N = 308$, $D = 6$).
Each cell uses $10$ random $90{:}10$ train/test splits.
Inputs and targets are standardised to zero mean and unit variance using training-set statistics.

\paragraph{Optimisation protocol (HSGP, VFF, VISH)}
For HSGP, VFF, and VISH, each cell is fit with a single L-BFGS-B run over $(\log \ell, \log \sigma^2_{\mathrm{sig}}, \log \sigma^2_{\mathrm{noise}})$ from the unit starting point $\theta_0 = (\log \ell_d = 0, \log \sigma^2_{\mathrm{sig}} = 0, \log \sigma^2_{\mathrm{noise}} = \log 0.1)$.
The same starting point is used for selection and for the subsequent kernel-hyperparameter fit.
We do not multi-start.
Variability across the $10$ splits is reported in every per-family figure as the median test NLL with shaded interquartile bands, robust to occasional optimisation outliers.

\paragraph{HSGP setup}
For HSGP, we use a Mat\'ern-$5/2$ ARD kernel on a box $[-L_d, L_d]^D$ per axis with $L_d = 1.2 \cdot \max_i |x_{id}|$.
The candidate dictionary is $\{1, \ldots, M_{\text{per-dim}}\}^D$, with $M_{\text{per-dim}}$ chosen so that the total candidate count satisfies $J \le 8000$.
The fit objective is the closed-form marginal likelihood of the truncated linear-Gaussian model described in Appendix~\ref{app:hsgp-detail}.
The non-uniform truncation baseline is constructed as documented in Appendix~\ref{app:hsgp-detail}.

\paragraph{VFF setup}
For VFF, we use the per-axis Mat\'ern-$5/2$ $K_{uu}$ blocks of \citet{Hensman2017}.
The candidate dictionary is $\{(d, j_d) : 1 \le d \le D,\, 0 \le j_d < M_{\text{per-dim}}\}$.
It is treated as a flat set and ranked globally.
The fit objective is the closed-form collapsed Titsias bound described in Appendix~\ref{app:vff-detail}, and the truncation baseline is the one documented there.

\paragraph{VISH setup}
For VISH, inputs are projected to $S^{D}$ via $z_n = (x_n, 1)/\|(x_n, 1)\|$ \citep{Dutordoir2020}.
Spherical harmonics and the order-$1$ arc-cosine Funk--Hecke spectrum are taken from the gpfy library \citep{Eleftheriadis2023, Cho2009}, with $\ell_{\max}$ chosen per $\bar D$ to fit gpfy's pre-computed fundamental-set tables.
The fit objective is the collapsed Titsias bound on the gpfy spherical-harmonic basis described in Appendix~\ref{app:vish-detail}.
Cumulative-shell truncation cuts at the largest degree $L^\star$ satisfying $\sum_{\ell \le L^\star} N(\bar D, \ell) \le M$.
The Eleftheriadis phase-truncation baseline uses gpfy's \texttt{phase\_truncation} parameter, choosing the per-shell phase count $m^\star$ so the total basis-function count is closest to the target $M$.

\clearpage
\section{Test RMSE on UCI for Experiments I, II, III}
\label{app:rmse-scores}

\begin{figure}[!htbp]
  \centering
  \resizebox{\linewidth}{!}{\import{figures/}{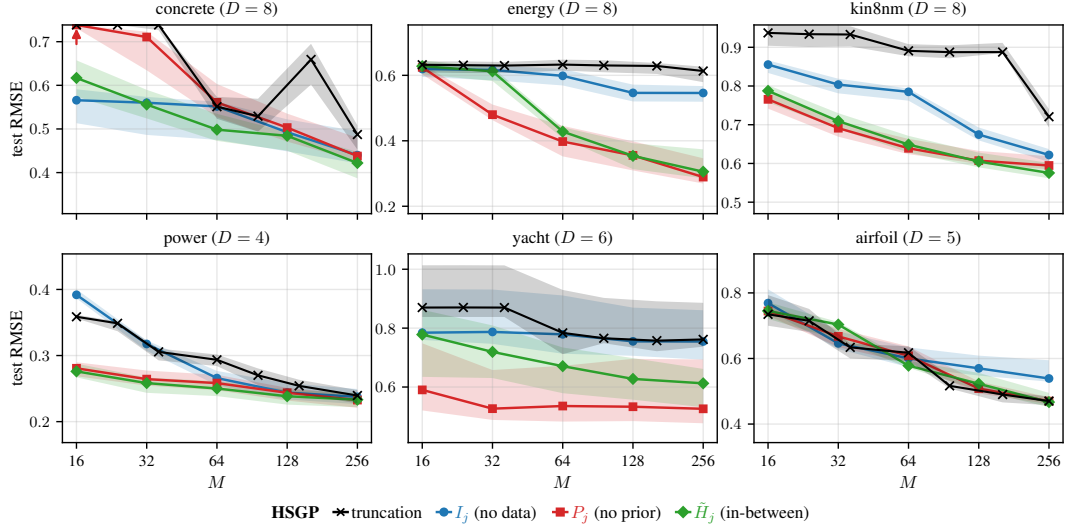}}
  \vspace{-0.8em}
  \caption{Experiment I: HSGP test RMSE on the same six UCI benchmarks and budgets as Figure~\ref{fig:hsgp-uci}.}
  \label{fig:app-hsgp-rmse}
\end{figure}

\begin{figure}[!htbp]
  \centering
  \resizebox{\linewidth}{!}{\import{figures/}{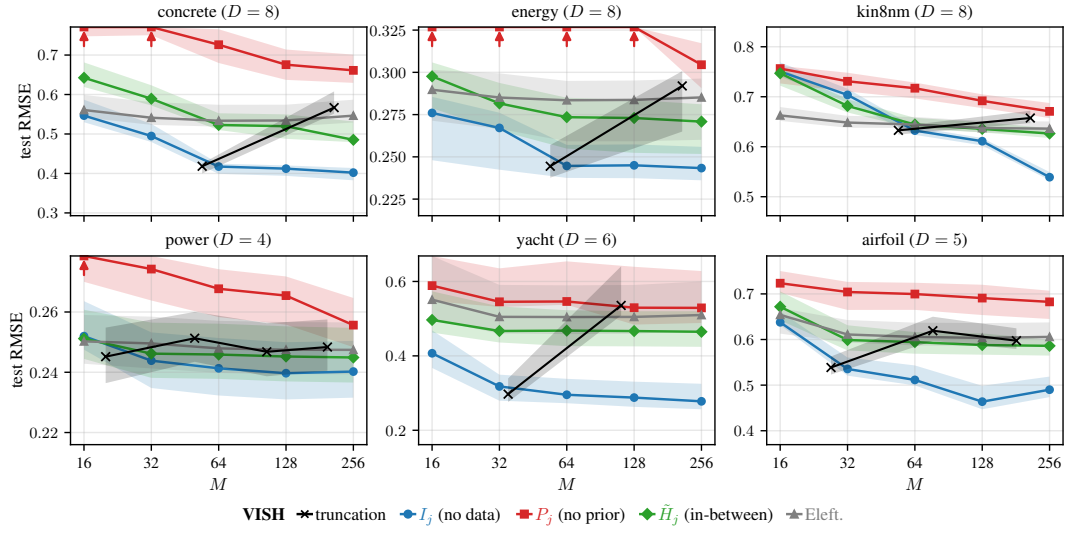}}
  \vspace{-0.8em}
  \caption{Experiment III: VISH test RMSE on the same six UCI benchmarks and budgets as Figure~\ref{fig:vish-uci}.}
  \label{fig:app-vish-rmse}
\end{figure}

Figures~\ref{fig:app-hsgp-rmse}, \ref{fig:app-vff-rmse}, and~\ref{fig:app-vish-rmse} repeat the UCI comparison with median test root mean square error (RMSE) versus budget $M$ across the same $10$ train/test splits, with interquartile bands.
The pattern across the three families is the same as the test-NLL pattern in the main text.
On HSGP the data-aware criteria $\colH{\tilde H_j}$ and $\colP{P_j}$ improve on $\colI{I_j}$ on most cells.
On VFF the three criteria are roughly tied and the per-axis truncation baseline tracks $\colI{I_j}$ closely.
On VISH the (no data) $\colI{I_j}$ matches or exceeds the data-aware criteria, and the cumulative-shell truncation curve shows the same odd-shell jumps as the test-NLL curve.
We use NLL as the headline metric in the body for two reasons: it captures the predictive distribution rather than only the point prediction, and it is the standard test metric for sparse GP regression in this literature.

\begin{figure}[!htbp]
  \centering
  \resizebox{\linewidth}{!}{\import{figures/}{lbfgs_vff_per_family_rmse.pgf}}
  \vspace{-0.8em}
  \caption{Experiment II: VFF test RMSE on the same six UCI benchmarks and budgets as Figure~\ref{fig:vff-uci}.}
  \label{fig:app-vff-rmse}
\end{figure}

\clearpage

\end{document}